\documentclass[letterpaper]{article} 
\usepackage{aaai24}  
\usepackage{times}  
\usepackage{helvet}  
\usepackage{courier}  
\usepackage[hyphens]{url}  
\usepackage{graphicx} 
\usepackage{natbib}  
\usepackage{caption} 
\usepackage{amsmath}
\usepackage{arydshln}
\usepackage{multirow}
\usepackage{graphicx}
\usepackage{tablefootnote}
\usepackage{algorithm}
\usepackage{amsthm,amsmath,amssymb}
\usepackage{mathrsfs}
\usepackage{algorithmic}

\usepackage{newfloat}
\usepackage{listings}
\DeclareCaptionStyle{ruled}{labelfont=normalfont,labelsep=colon,strut=off} 
\floatstyle{ruled}
\newfloat{listing}{tb}{lst}{}
\floatname{listing}{Listing}
\usepackage{hyperref}

\title{TA\&AT: Enhancing Task-Oriented Dialog with Turn-Level Auxiliary Tasks and Action-Tree Based Scheduled Sampling}
\author{
    Longxiang Liu\textsuperscript{\rm 1,2}, Xiuxing Li\textsuperscript{\rm 1,2}, Yang Feng\textsuperscript{\rm 1,2,\thanks{Corresponding author.}}
   \\
}
\affiliations{
    \textsuperscript{\rm 1}Key Laboratory of Intelligent Information Processing,\\ Institute of Computing Technology, Chinese Academy of Sciences (ICT/CAS)\\
    \textsuperscript{\rm 2}University of Chinese Academy of Sciences, Beijing, China\\
    \{liulongxiang21s, lixiuxing, fengyang\}@ict.ac.cn
}
\usepackage{bibentry}

\begin{document}

\maketitle

\begin{abstract}
Task-oriented dialog systems have witnessed substantial progress due to conversational pre-training techniques. Yet, two significant challenges persist. First, most systems primarily utilize the latest turn's state label for the generator. This practice overlooks the comprehensive value of state labels in boosting the model's understanding for future generations. Second, an overreliance on generated policy often leads to error accumulation, resulting in suboptimal responses when adhering to incorrect actions. To combat these challenges, we propose turn-level multi-task objectives for the encoder. With the guidance of essential information from labeled intermediate states, we establish a more robust representation for both understanding and generation. For the decoder, we introduce an action tree-based scheduled sampling technique. Specifically, we model the hierarchical policy as trees and utilize the similarity between trees to sample negative policy based on scheduled sampling, hoping the model to generate invariant responses under perturbations. This method simulates potential pitfalls by sampling similar negative policy, bridging the gap between task-oriented dialog training and inference. Among methods without continual pre-training, our approach achieved state-of-the-art (SOTA) performance on the MultiWOZ dataset series and was also competitive with pre-trained SOTA methods.
\end{abstract}

\section{Introduction}\label{sec:intro}
\begin{figure}
\centering
\includegraphics[width=0.9\columnwidth]{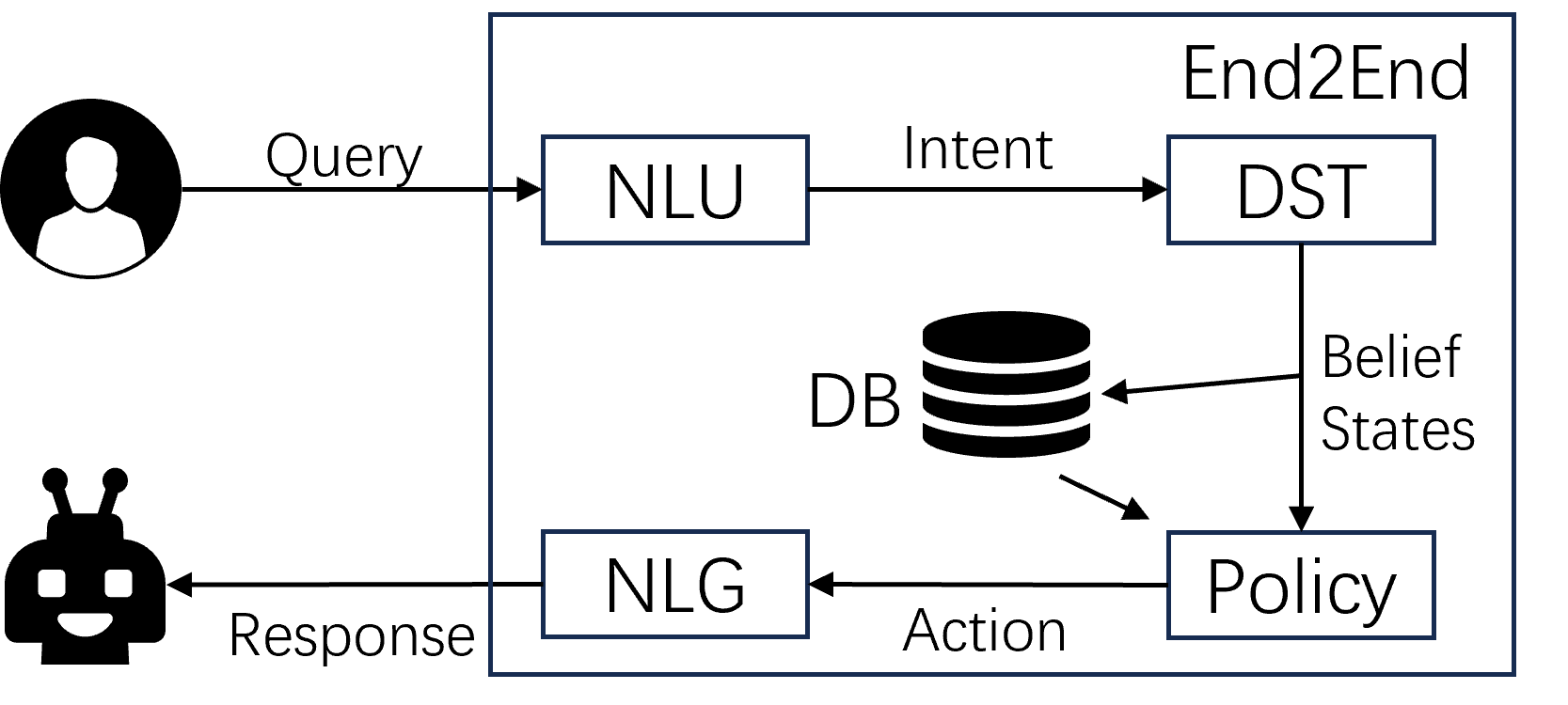} 
\caption{Illustration of task-oriented dialog system.}
\label{fig:tod}
\end{figure}

The goal of task-oriented dialog (TOD) is to better accomplish a user-specific task through multi-turn dialog. As shown in Figure \ref{fig:tod}, a typical TOD system consists of four modules: (1) natural language understanding (NLU) to determine the user intent. (2) dialog state tracking (DST) to extract the user constraints which will be used to query the database (DB). (3) policy (POL) to plan for the system's next action sequence. (4) natural language generation (NLG) to generate a fluent and informative response. In recent works, NLU is usually not handled specifically, but put into DST module~\cite{takanobu2020your}. End-to-end task-oriented dialog, which is also the focus of our work, integrates submodules into one model for joint training.

Performance of end-to-end TOD systems has improved dramatically in recent years thanks to powerful pre-trained language models, especially dialog pre-training. However, two issues still exist. \textbf{Firstly}, there are some datasets with intermediate state annotations, which most works simply use to supervise the generator. While we believe that the annotations are not fully utilized and their essential value for understanding is overlooked. \textbf{Secondly}, the sequence-to-sequence (Seq2Seq) training approach leads to error accumulation, especially generating unsatisfying responses that are attached to incorrect actions.

To solve the first problem, we utilize the labels of intermediate states to supervise the hidden states output by encoder, hoping better representations provide useful clues for the subsequent generation. Inspired by MTTOD~\cite{lee2021improving}, which utilizes the belief state annotations to construct a context-span labeling auxiliary task, we leverage more annotations to construct more \textbf{auxiliary tasks} (e.g., slot type, slot change, action type, and response keywords prediction). Besides, inspired by DialoFlow~\cite{li2021conversations}, we optimize the \textbf{turn-level} representation instead of token-level since it reflects higher-level information such as conversational goal or potential influence before generation of next response.

To solve the second problem, we attempt to use \textbf{scheduled sampling} technique~\cite{bengio2015scheduled} to reduce the inconsistency between training and inference. However, in TOD system, simply using the token-level scheduled sampling does not actually simulate the errors at inference. Given a specific token, the likelihood of generating a subsequent token is highly deterministic due to the strong conditional relationships between tokens. This results in a sharp token-level conditional probability distribution, making it challenging for a single negative token to be sampled. Instead, there is more uncertainty among action sequences. Therefore, we propose a method that can directly sample a negative action sequence similar to the ground truth action at the training time, called \textbf{action-tree based scheduled sampling}. Specifically, inspired by SPACE~\cite{he2022unified}, we model the action sequence as a tree, calculate the similarity according to the edit distance between action trees, and then use similarity as the sampling distribution of negative action sequences. We optimize the likelihood of reference response under the perturbation of action sequence.

We have conducted comprehensive experiments on MultiWOZ 2.0/2.1/2.2. Experiments show that our method \textbf{TA\&AT} substantially improves TOD system and achieves new state-of-the-art results among methods that \textbf{do not adopt continual pre-training}, pushing the end-to-end combined score on MultiWOZ 2.0/2.1/2.2 to \textbf{109.27/108.03/103.59}. Ablation study also verifies the effectiveness of our proposed method.

In summary, our main contributions are three-fold:
\begin{itemize}
    \item We explore how to make the most of intermediate annotations in TOD system, through turn-level auxiliary tasks. 
    \item To the best of our knowledge, this is the first attempt to introduce sequence-level scheduled sampling into TOD.
    \item Extensive experiments show our method achieves state-of-the-art performance on MultiWOZ datasets.
\end{itemize}

\section{Related Work}
End-to-end task-oriented dialog aims at jointly training submodules and building a text-in, text-out integrated system. \cite{wen2016network} first proposed a trainable neural network-based framework for end-to-end TOD, using CNN \cite{kalchbrenner2014convolutional} and LSTM \cite{hochreiter1997long} in different modules. \cite{lei2018sequicity, zhang2020probabilistic, zhang2020task} proposed their methods mainly based on CopyNet \cite{gu2016incorporating}  in seq2seq training and elaborate design of decoder. 

Due to the blooming of pre-trained language models (PLMs), recent approaches employ PLM as their backbone such as GPT \cite{radford2018improving}, T5 \cite{raffel2020exploring} and UniLM \cite{dong2019unified}. \cite{kulhanek2021augpt, peng2021soloist, yang2021ubar, hosseini2020simple} applied GPT-2 model for different modules, training in turn-level or session-level. Since there are not only generation tasks but also language understanding tasks in TOD, encoder-decoder framework fits better. There are many works that use T5 as a base model and promote end-to-end performance from their own perspectives. Among them, \cite{su2021multi, lee2021improving, bang2023task} utilize multi-task learning, \cite{sun2023mars} leverages contrastive learning to model the relationship between dialog context and belief/action state representations. There are also works based on parameter-shared encoder-decoder UniLM, \cite{he2022galaxy, he2022unified} continually pre-train their proposed semi-supervised or self-supervised learning tasks on UniLM and then adapt to downstream tasks through finetuning, which achieves current state-of-the-art.

To mitigate error accumulation in end-to-end TOD, \cite{zhang2020task} takes different valid dialog policies into consideration to learn a balanced action distribution, guiding the dialog model to generate diverse responses. \cite{sun2022bort} introduced a back and denoising reconstruction approach and \cite{he2022galaxy} employed consistency regularization to refine the learned representation. Different from above works, we attempted to apply scheduled sampling, which is proposed in sequence generation task \cite{bengio2015scheduled} and improved in neural machine translation \cite{zhang2019bridging}.

\section{Model Framework}\label{sec:framework}
In this section, we will introduce our model framework. As described in Section \ref{sec:intro}, end-to-end task-oriented dialog generation is usually modeled as a cascading generation problem. In each turn, the system receives the user's input, which will be concatenated with the context information in the memory block. Then the belief states should be generated, which is a hierarchical semantic state reflecting the constraints of user requests. The belief states are used to query the database, whose matching results will be used together with the context information to determine a policy. Policy is a hierarchical action sequence, guiding the process of response generation. In general, belief states contain \textit{(domain, slot, value)} and policy contains \textit{(domain, action, slot)}, both of which are three-level structures. 

\begin{figure}[!htb]
\centering
\includegraphics[width=\columnwidth]{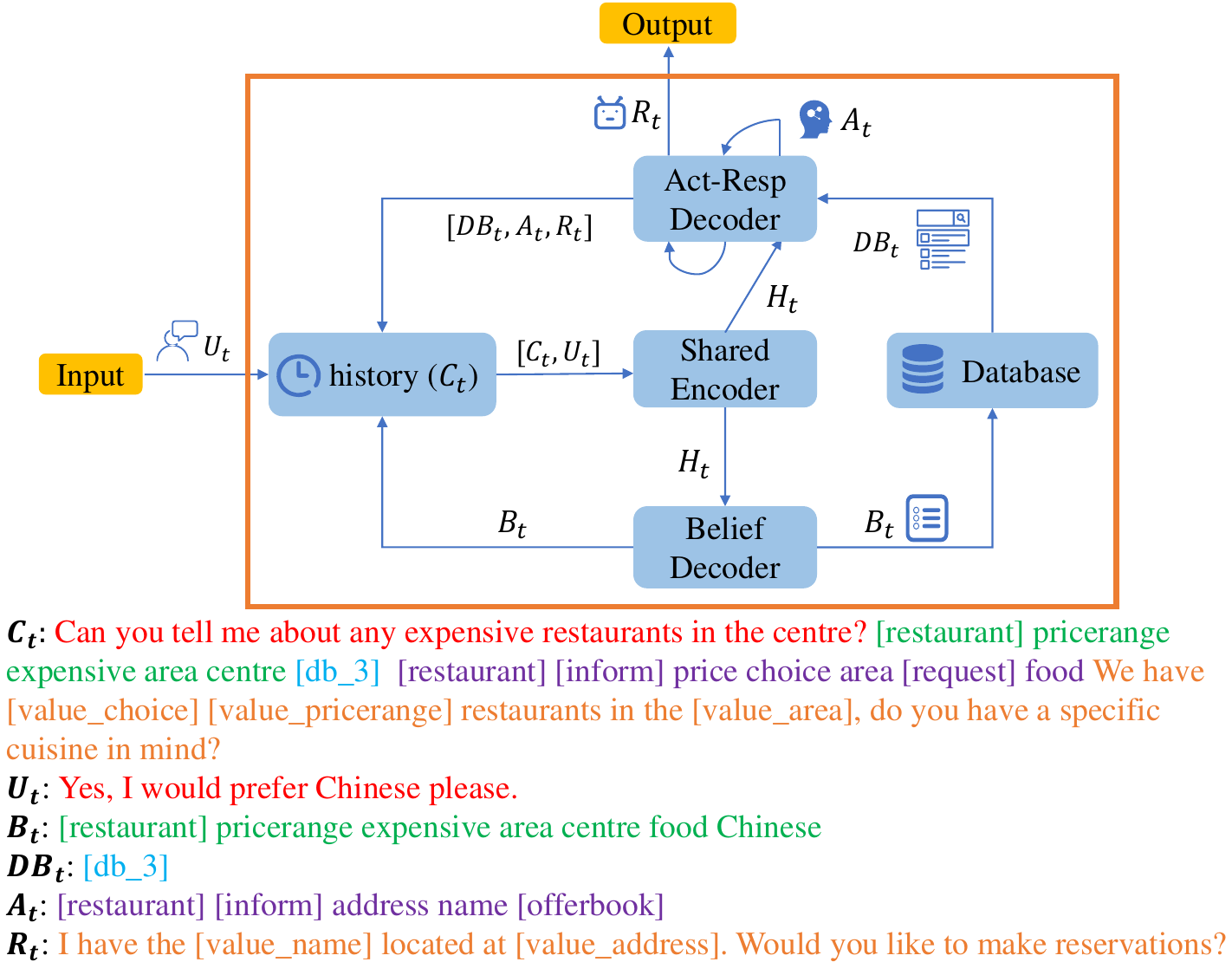} 
\caption{Illustration of our task-oriented dialog system framework. For simplicity, we show an example dialog in the scenario of a user ordering a restaurant, $t=1$ (starts from 0). The memory module will keep track of the new generated belief states, db states, acts, and responses.}
\label{fig:framework}
\end{figure}

There are several choices for the base model framework, like decoder-only GPT \cite{yang2021ubar, peng2021soloist}, encoder-decoder T5 \cite{su2021multi, bang2023task}, UniLM-based models \cite{he2022galaxy, he2022unified}, encoder-2decoders based models \cite{lee2021improving, cholakov2022efficient}. Considering that the belief generation depends more on understanding and summarization ability, while the policy and response generation relies more on generative ability to maintain contextual coherence. We believe that they belong to different semantic subspace, and in our experiments UniLM requires time-consuming pre-training to show good performance, which is also verified in \cite{he2022galaxy}. We finally adopt the framework proposed in \cite{lee2021improving}, containing one shared encoder and two different decoders, as shown in Figure \ref{fig:framework}.

\subsection{Definitions}
Here we introduce the symbols involved according to the input stream. In the $t$-th turn of a dialog, $U_t$ represents the user input utterance. $B_t$ is the belief state, which in the Figure \ref{fig:framework} is \textit{\{restaurant:\{pricerange:expensive, area:centre, food:Chinese\}\}}. $DB_t$ represents the database result, reflecting the matching number of entities satisfying the belief states. $A_t$ represents the action sequence, which in the Figure \ref{fig:framework} is \textit{\{restaurant:\{inform:[address,name], offerbook:[]\}\}}. $R_t$ represents the system response. The context information $I_t=(U_t, B_t, DB_t, A_t, R_t)$ will be gathered in the memory block. Note that inspired by \cite{yang2021ubar}, we concatenate all the history information, where $C_t = \text{Concat(}I_0,...I_{t-1}\text{)}$. 
\subsection{Objectives}
In the end-to-end task-oriented dialog framework, the context information in memory module and current user utterance will be input to a shared transformer encoder to get the hidden states $H_t$. Then $H_t$ is first input to the belief decoder to generate belief states. The generated belief states $B_t$ will be used to query the database, returning $DB_t$. Finally, $H_t$ and $DB_t$ are input together to the Action-Response Decoder to autoregressively generate the action $A_t$ and response $R_t$. 

\begin{small}
\begin{equation}
    \label{eq:encdec}
    \begin{split}
        H_t &= \text{Encoder}([C_t, U_t]) \\
        B_t &= \text{Decoder}_{b}(H_t) \\
        A_t, R_t &= \text{Decoder}_{ar}(H_t, DB_t)
    \end{split}
\end{equation}
\end{small}

Both decoders and the encoder are optimized with cross entropy loss supervised by the teacher-forcing ground truth belief states, action, and response. 
\begin{small}
\begin{equation*}
    \begin{split}
        \mathcal{L}_B &= -\log P(\hat{B}_t|H_t) \\
        \mathcal{L}_{AR} &= -\log P(\hat{A_t}, \hat{R_t}|H_t, DB_t) \\
        \mathcal{L} &= \mathcal{L}_B + \mathcal{L}_{AR} \\
    \end{split}
\end{equation*}
\end{small}
\section{Methodology}
\begin{figure*}[!htb]
\centering
\includegraphics[width=2\columnwidth]{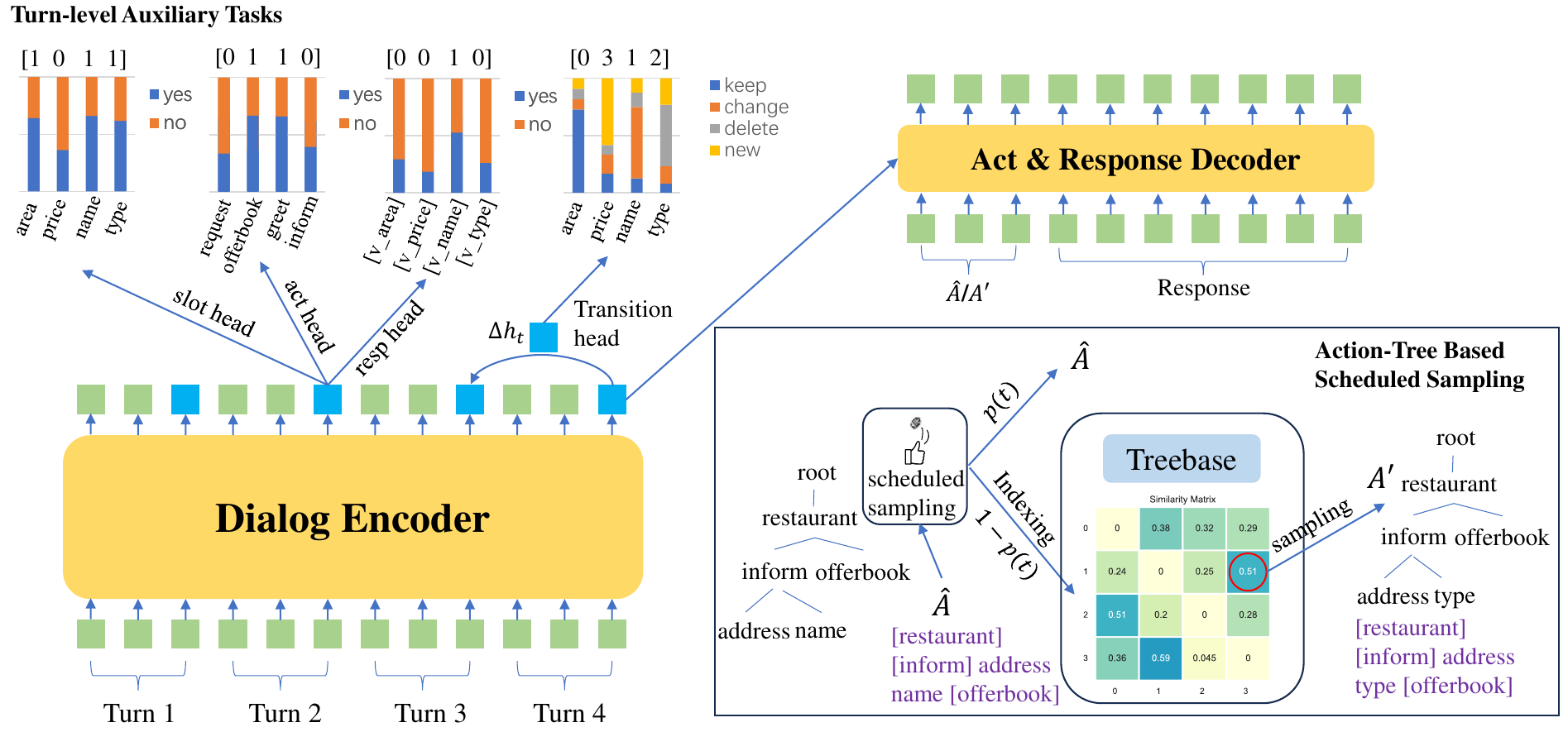} 
\caption{Overall framework of our proposed methods. The left part shows the process of extracting turn-level representations and passing them to four multi-dimensional Bernoulli/Categorical classification heads. The right part shows the process of action-tree based scheduled sampling, where the ground truth action $\hat{A}$ will be replaced with the probability of $1-p(t)$, a replacing action sample $A'$ is then sampled according to the normalized similarity score. The calculation of similarity score is based on the action-tree Editing Distance, which will be discussed detailedly
 in Section \ref{sec:at}.}
\label{fig:overall}
\end{figure*}

In this section, we elaborate on our proposed method. In order to relieve the problem of \textbf{insufficient} utilization of labels we described in Section \ref{sec:intro}, we propose four turn-level auxiliary tasks to enhance the understanding ability of encoder, providing some inherent clues for subsequent generation; To alleviate the problem of \textbf{sequence-level error accumulation}, we propose action-tree based scheduled sampling, making response generation more robust. We will first describe the auxiliary tasks, then describe the action-tree based scheduled sampling approach, and discuss the training and inference process at last. Our proposed method is shown in Figure \ref{fig:overall}.

\subsection{Turn-Level Auxiliary Tasks}
There are many annotations other than ground truth responses, which can be used to strengthen the understanding of the encoder. Inspired by MTTOD \cite{lee2021improving}, which leverages the annotations of belief states to introduce a simple span prediction task for task-oriented dialog enhancement, we propose to leverage more types of annotations and introduce additional auxiliary tasks. Besides, as stated in DialoFlow \cite{li2021conversations}, the turn-level representations reflect higher-level information such as the conversational goal or potential influence before generation of next responses. Based on the above two points, we supervise turn-level representation learning by using high-level supervision signals from different types of annotations (e.g., belief states, actions, responses). We hope that these turn-level representations can better provide clues for subsequent generation. Four below auxiliary tasks are proposed, and each corresponding \textbf{true} label set in the example of Figure \ref{fig:framework} is given behind. For simplicity, domain labels are ignored.
\begin{itemize}
    \item Slot type: \textit{[pricerange, area, food]}
    \item Slot transition: \textit{\{pricerange:keep, area:keep, food:new\}}
    \item Action type: \textit{[inform, offerbook]}
    \item Response keywords: \textit{([value\_name],[value\_address])}
\end{itemize}
Now we describe these four tasks in detail.
\subsubsection{Turn representation}
Both our model encoder and two decoders are initialized using T5's corresponding modules, which is not the focus of this paper, so the introduction of T5's model structure is ignored. According to Equation \ref{eq:encdec}, $H_t$ is the encoder output hidden states, we use the end position of each turn to select the turn-level representations from turn-0 to turn-$(t-1)$, denoted as $T_t$.
\begin{small}
\begin{equation*}
    \begin{split}
         P^{end}_t &= [pos_0, pos_1, ..., pos_{t-1}] \in \mathbb{N}^t \\
         T_t &= \text{IndexSelect}(H_t, P^{end}_t) \in \mathbb{R}^{d \times t}
    \end{split}
\end{equation*}
\end{small}

\subsubsection{Slot Type Prediction}
Determining which slots are mentioned in a user's utterance can help to generate belief states since it narrows the scope of slot-value pairs, and such discriminative task is better suited to the capabilities of the encoder. There are some turns associated with multiple types of slots mentioned. Following GALAXY \cite{he2022galaxy}, we model the slot type prediction task as a multi-label classification problem. In Equation \ref{eq:slottype}, we denote $ST = (st_1, st_2, ..., st_N)$, where N is the total number of slot types. A multi-dimensional Bernoulli distribution is used for modeling the slot types. The turn representation $T$ will be passed through a multi-dimensional binary classifiers to get the prediction score of each slot type.
\begin{small}
\begin{equation}
    \begin{split}
     \label{eq:slottype}
        &p(ST|T)=\prod \limits_i^Np(st_i|T) \\
        &p(st_i|T)=\text{sigmoid} (W_{st}T)\in \mathbb{R}^N \\
        \mathcal{L}_{st}=-\sum_{i=1}^N\{y_i&\log p(st_i|T) + (1-y_i)\log (1-p(st_i|T))\}
    \end{split}
\end{equation}
\end{small}

where $W_{st}$ is trainable parameter matrix of linear slot type head and $y_i \in \{0,1\}$ is the label of whether $st_i$ appears in current turn.
\subsubsection{Slot Change Prediction}
SOM-DST \cite{kim2019efficient} has mentioned that state operation prediction allows state tracking model to efficiently generate the values of only a minimal subset of the slots. In our situation, predicting the slot change also provides important clues for belief state generation. We define slot change into four categories: \{keep, change, delete, new\}, which is similar to the operations in database system. Given two consecutive turns' belief states, the slot change between them is easy to get. Here a multi-dimensional categorical distribution is adopted for modeling the slot change, as shown in Equation \ref{eq:slotchange}, we denote $SC = (sc_1, sc_2, ..., sc_N)$. $\Delta T = T_{t} - T_{t-1}$, which reflects the difference between adjacent turns' representations, is fed to the trainable transition head $W_{sc}$. Note that for simplicity, the subscript of $T_t$ is omitted when there is no ambiguity. 
\begin{small}
\begin{equation}
    \begin{split}
     \label{eq:slotchange}
        \Delta T_t &= T_{t} - T_{t-1} \\
        p(SC|\Delta T) &=\prod \limits_i^{|SC|}p(sc_i^{y_i}|\Delta T) \\
        p(sc_i|\Delta T)&=\text{Softmax} (W_{sc}\Delta T)\in \mathbb{R}^{4} \\
        \mathcal{L}_{sc}=-&\sum_{i=1}^N \log p(sc_i^{y_i}|\Delta T)
    \end{split}
\end{equation}
\end{small}

where $y_i \in \{0,1,2,3\}$ is the label of $i$-th slot change.
\subsubsection{Action Prediction}
As described in GALAXY \cite{he2022galaxy}, identifying the actions (e.g. request, offerbook, etc.) can facilitate learning better representations for policy optimization to improve the overall end-to-end performance. Here we adopt the same way in GALAXY using multi-dimensional Bernoulli distribution to model action prediction. The difference is that we predict all turns' actions while GALAXY only predicts current turn's. 
\begin{small}
    \begin{equation}
    \begin{split}
     \label{eq:acttype}
        &p(A|T)=\prod \limits_i^N p(a_i|T) \\
        &p(a_i|T)=\text{sigmoid} (W_{a}T)\in \mathbb{R}^N \\
        \mathcal{L}_{a}=-\sum_{i=1}^N\{y_i&\log p(a_i|T) + (1-y_i)\log (1-p(a_i|T))\}
    \end{split}
\end{equation}
\end{small}

where $W_{a}$ is trainable parameter matrix of linear action head and $y_i \in \{0,1\}$ is the label of whether $a_i$ is taken in current action.

\subsubsection{Response Keywords Prediction}
In most situations in task-oriented dialog, the system should inform some essential values in the response according to what the user requests, which also relates to the evaluation metric \textbf{Success} in Section \ref{sec:eval}. Predicting such keywords can make the model focus on essential information to be generated, such as \textit{[value\_name], [value\_area]}, etc. In delexicalized responses \cite{zhang2020task}, words like \textit{[value\_xxx]} are in a finite set. Here we model the bag-of-words predictions as a multi-dimensinal Bernoulli distribution, as shown in Equation \ref{eq:resp}.
\begin{small}
    \begin{equation}
    \begin{split}
     \label{eq:resp}
        &p(K|T)=\prod \limits_i^N p(k_i|T) \\
        &p(k_i|T)=\text{sigmoid} (W_{k}T)\in \mathbb{R}^N \\
        \mathcal{L}_{k}=-\sum_{i=1}^N\{y_i&\log p(k_i|T) + (1-y_i)\log (1-p(k_i|T))\}
    \end{split}
\end{equation}
\end{small}

where $N$ is the vocabulary size of keywords, $W_{k}$ is trainable parameter matrix of linear response head and $y_i \in \{0,1\}$ is the label of whether keyword $k_i$ is appeared in the response.

To be summarized, total loss for the turn-level auxiliary tasks is
\begin{equation*}
    \mathcal{L}_{TA} = \mathcal{L}_{st} + \mathcal{L}_{sc} + \mathcal{L}_{a} + \mathcal{L}_{k}
\end{equation*}
\subsection{Action-Tree Based Scheduled Sampling}
\label{sec:at}
In our experiments, we found that the more we train, the more faithful the generated responses are to the generated action sequence. However, this can lead to unsatisfying responses, especially when the generated actions are not reasonable enough. This phenomenon is called error accumulation, which is caused by exposure bias. Since the seq2seq training process is teacher-forced, it is inconsistent with the inference phase \cite{zhang2019bridging}. Scheduled sampling \cite{bengio2015scheduled} is a straightforward way to mitigate this problem, which randomly replaces target-side input tokens with model predictions following a curriculum learning strategy. 

However, directly adopting token-level scheduled sampling is not effective for our task, because the main error at inference exists in the action sequence bringing inaccurate response sequence, so sequence-level replacement is needed. To this end, we propose an action-tree based scheduled sampling method. Next we describe the method in detail.
\subsubsection{Action Tree}
Inspired by SPACE \cite{he2022unified}, we calculate the similarity score among action sequences and save the similarity matrix. When calculating the similarity score, we first convert the action sequence to a hierarchical action tree, containing the tertiary structure (domain,action,slot) from top to bottom, as shown in the right part of Figure \ref{fig:overall}. Then we derive the Tree Editing Distance \cite{zhang1989simple}, which is the weighted number of edit operations (insert, delete, and modify) to transform one tree to another. Note that we use \textbf{ordered tree}, which is different from SPACE, since in our experiments we found the relative position of actions will affect the response generation. Besides, reordering to an unordered tree may result in some not existing action sequences. 
Denoting the semantic trees of $i$-th action and $j$-th action are $T_i$ and $T_j$. Tree Editing Distance is calculated, and then similarity score $s_{i,j}$ between $i$-th action and $j$-th action is calculated by Equation \ref{eq:tree}.
\begin{small}
    \begin{equation}
    \begin{split}
        \label{eq:tree}
        s_{i,j} &= \dfrac{\max \{|T_i|, |T_j|\} - d_{i,j}}{\max \{|T_i|, |T_j|\}} \\
        d_{i,j} &= \text{TreeEditingDistance}(T_i,T_j)
    \end{split}
\end{equation}
\end{small}
\subsubsection{Scheduled Sampling}
As shown in the bottom right of Figure \ref{fig:overall}, in the training process, before one ground truth action $\hat{A}_t$ is input to the act-response decoder, it will be retained with probability $p(t)$, which is calculated by Equation \ref{eq:ss} \cite{zhang2019bridging}.
\begin{small}
    \begin{equation}
\label{eq:ss}
    p = \dfrac{\mu}{\mu + \exp (t/\mu)}
\end{equation}

where $\mu$ is a hyper-parameter. And the function is strictly monotone decreasing. 
\end{small}
Otherwise, the ground truth action will be used to index the similarity matrix, assuming the indexed column is $i$. We denote the similarity matrix as $M$, then the sampling distribution is
\begin{small}
    \begin{equation*}
    p_j^* = \dfrac{M[i,j]}{\sum_{j=1, j\ne i}^N M[i,j]}
\end{equation*}
\end{small}

note that here we guarantee that the same $i$-th action will not be sampled. 

\subsubsection{Loss}
As shown in Equation \ref{eq:loss}, when the perturbed action is adopted as input, the action loss should not be optimized but the response loss should still be optimized to improve the robustness of response generation in the presence of noisy actions. In such situation, model should learn to depend more on the context when generating the response.
\begin{small}
    \begin{equation}
    \label{eq:loss}
    \begin{split}
        \mathcal{L}_A = -\log P(A_t|H_t,DB_t) \\
        \mathcal{L}_R = -\log P(R_t|H_t,DB_t,A_t) \\
        \mathcal{L}_{AT} = \begin{cases}
    	\mathcal{L}_A+\mathcal{L}_R, &A_t=\hat{A_t}\\
    	\mathcal{L}_R, &A_t=A'_t
    		   \end{cases}
    \end{split}
\end{equation}
\end{small}

\subsection{Traning and Inference}
The final loss in our training process is described by Equation \ref{eq:total}.
\begin{equation}
\label{eq:total}
    \mathcal{L} = \mathcal{L}_{TA}+\mathcal{L}_{B}+\mathcal{L}_{AT}
\end{equation}

Note that since the belief decoder is not our focus in this work, we did not discuss this module and omit it in Figure \ref{fig:overall}, but the loss $\mathcal{L}_{B}$ always exists.

In the inference phase, we only utilize the shared encoder and two decoders, and neither the classification head nor scheduled sampling is required, making the overall inference cost completely unchanged with respect to our backbone.
\section{Experiments}
In this section, we will introduce experimental data, metrics, compared baselines, and our results in different tasks. Our code is released in our github repository\footnote{\url{https://github.com/ictnlp/TA-AT}}.
\begin{table*}
\large
\centering
\resizebox{2\columnwidth}{!}{%
\begin{tabular}{l|cccc|cccc|cccc}
\hline
\multirow{2}{*}{Model} & \multicolumn{4}{c|}{MultiWOZ 2.0} & \multicolumn{4}{c|}{MultiWOZ 2.1} & \multicolumn{4}{c}{MultiWOZ 2.2} \\ 
          & Inform & Success & BLEU  & Comb   & Inform & Success & BLEU  & Comb   & Inform & Success & BLEU  & Comb  \\ \cline{1-13}
\hline
\multicolumn{13}{l}{\textit{w.o. continual pre-training}}\\
\hline
SimpleTOD & 84.40   & 70.10    & 15.01 & 92.26  & 85.00     & 70.50    & 15.23 & 92.98  & -      & -       & -     & -     \\
DoTS      & 86.59  & 74.14   & 15.06 & 95.43  & 86.65  & 74.18   & 15.90  & 96.32  & 80.40   & 68.70    & 16.80  & 91.40  \\
SOLOIST  & 85.50   & 72.90    & 16.54 & 95.74  & -      & -       & -     & -      & 82.30   & 72.40    & 13.60  & 90.9  \\
MinTL     & 84.88  & 74.91   & 17.89 & 97.79  & -      & -       & -     & -      & 73.70   & 65.40    & 19.40  & 89.00    \\
UBAR      & \textbf{95.40}   & 80.70    & 17.00    & 105.05 & \textbf{95.70}   & 81.80    & 16.50  & 105.25 & 83.40   & 70.30    & 17.60  & 94.40  \\
GALAXY    & 93.10   & 81.00      & 18.44 & 105.49 & 93.50   & 81.70    & 18.32 & 105.92 & 85.40   & 75.70    & 19.64 & 100.20 \\
BORT      & 93.80   & \textbf{85.80}    & 18.50  & 108.30  & -      & -       & -     & -      & 85.50   & 77.40    & 17.90  & 99.40  \\
Mars      & -      & -       & -     & -      & -      & -       & -     & -      & \textbf{89.20}   & \textbf{80.30}    & 19.00    & 103.40 \\
MTTOD     & 90.99  & 82.58   & 20.25 & 107.04 & 90.99  & 82.08   & 19.68 & 106.22 & 85.90   & 76.50    & 19.00    & 100.20 \\
\hdashline
\textbf{TA\&AT}    & 93.60   & 83.60    & \textbf{20.67} & \textbf{109.27} & 92.50   & \textbf{84.00}    & \textbf{19.78} & \textbf{108.03} &   86.40     &   80.10  &  \textbf{20.34}  &  \textbf{103.59}   \\
\hline
\multicolumn{13}{l}{\textit{w. continual pre-training}}\\
\hline
PPTOD*     & 89.20   & 79.40    & 18.62 & 102.92 & 87.09  & 79.08   & 19.17 & 102.26 & 83.10   & 72.70    & 18.20  & 96.10  \\
GALAXY*    & 94.40   & 85.30    & 20.50  & 110.35 & 95.30   & 86.20    & 20.01 & 110.76 & -      & -       & -     & -     \\
SPACE*     & 95.30   & 88.00      & 19.30  & 110.95 & 95.60   & 86.10    & 19.91 & 110.76 & -      & -       & -     & -     \\ \hline
\end{tabular}%
}
\caption{E2E performances on MultiWOZ 2.0/2.1/2.2. TA\&AT is our method, short for Turn-level Auxiliary tasks and Action-Tree based scheduled sampling. All results are from original papers or public MultiWOZ leaderboard. `*' means using continual training on extra datasets.}
\label{tab:main}
\end{table*}

\subsection{Datasets and Evaluation Metrics}\label{sec:eval}
\subsubsection{Datasets}
We evaluate end-to-end dialog system performance of our proposed methods on public task-oriented dialog benchmark MultiWOZ \cite{budzianowski2018multiwoz}. We evaluate our method on MultiWOZ 2.0, 2.1 and 2.2. Following the data split in \cite{lee2021improving}, the number of train/validation/test set is 8438/1000/1000. And to reduce diversity of the surface form, we replace some specific slot values with \textit{[value\_xxx]} to construct the delexicalized response, allowing the model to learn value-independent parameters \cite{zhang2020task}.
\subsubsection{Metrics}
We follow the automatic evaluation metrics to evaluate the response quality for task-oriented dialog system on MultiWOZ datasets. \textbf{Inform rate} measures whether a dialog system has provided an accurate entity; \textbf{Success rate} measures whether a dialog system has answered all requested information; \textbf{BLEU} is computed with references, measuring the fluency of the generated response. \textbf{Combined score} = $(\text{Inform}+\text{Success})\times 0.5+\text{BLEU}$, reflects the overall quality of the dialog system, which is our main metric. 
\begin{table*}[!htb]
\centering
\resizebox{2\columnwidth}{!}{%
\begin{tabular}{l|cccc|cccc|cccc}
\hline
\multirow{2}{*}{Model} & \multicolumn{4}{c|}{10\% data}         & \multicolumn{4}{c|}{20\% data}         & \multicolumn{4}{c}{50\% data}          \\
                          & Inform & Success & BLEU  & Comb   & Inform & Success & BLEU  & Comb   & Inform & Success & BLEU  & Comb   \\ \hline
MinTL                  & 55.5   & 44.9    & 15.6  & 65.8   & 64.3   & 54.9    & 16.2  & 75.8   & 70.3   & 62.2    & 18.0    & 84.3   \\
PPTOD                  & 68.3   & 53.7    & 15.7  & 76.7   & 72.7   & 59.2    & 16.3  & 82.3   & 74.8   & 62.4    & 17.0    & 85.6   \\
UBAR                   & 50.3   & 34.2    & 13.5  & 55.8   & 65.5   & 48.7    & 14.5  & 71.6   & 77.6   & 63.3    & 16.3  & 86.8   \\
MTTOD                  & 66.9   & 55.2    & 13.8  & 74.9   & 75.0     & 63.3    & 14.3  & 83.5   & 78.5   & 67.5    & 15.2  & 88.2   \\
Mars                   & 69.4   & 55.3    & 15.6  & 78.0     & 76.7   & 62.9    & \textbf{17.2}  & 87.0     & 82.2   & 71.2    & \textbf{18.6}  & 95.3   \\ \hdashline
TA\&AT                 & \textbf{71.5}   & \textbf{58.4}      & \textbf{16.2} & \textbf{81.1} & \textbf{79.2}   & \textbf{68.2}    & 16.8 & \textbf{90.5} & \textbf{83.5}   & \textbf{73.8}    & 18.1 & \textbf{96.8} \\ \hline
\end{tabular}%
}
\caption{E2E results of low-resource experiments. 10\% (800 dialogs), 20\% (1600 dialogs), 50\% (4000 dialogs) of training data is used to train our model. All of the results are cited from Mars \cite{sun2023mars}.}
\label{tab:lowresource}
\end{table*}
\subsection{Settings}
Following \cite{lee2021improving}, we use a pre-trained T5-base model \cite{raffel2020exploring} to initialize our shared encoder and two decoders. We implement our methods based on the HuggingFace Transformers library \cite{wolf2020transformers}. We train our model for 10 epochs on a single 40G NVIDIA A100. Our model is trained for approximately 10 hours. In low resource setting, our model is trained for 20 epochs. The initial learning rate is set to 5e-4, batch size is set to 8 and the proportion of warmup steps is set to 0.1. We adopt an optimizer as AdamW \cite{loshchilov2017decoupled} with linear learning rate decay. We select the best model based on the performance on the validation set. For the hyperparameter $\mu$, we choose most suitable one from \{10,15,20\} for different datasets. To remove randomness, we fix our random seed to 42 in our experiments. A simple greedy search algorithm is used when decoding belief states, action, and responses.

\subsection{Baselines}
For a fair comparison, we confine our analysis to those methods that utilize PLMs. And the methods using PLMs typically fall under two distinct settings:
\begin{itemize}
\item \textbf{Without Continual}: Directly fine-tuning the PLM for specific downstream tasks, such as end-to-end modeling.
\item \textbf{With Continual}: Beginning with continual pre-training on extra datasets and then transitioning to fine-tuning.
\end{itemize}
We will compare our method with those in the \textbf{Without Continual} setting to underline the strengths of our approach. Additionally, comparisons with the methods in \textbf{With Continual} setting will be conducted to clearly illustrate the extent of the gap between our method and them. 

We compared serveral strong baselines, including SimpleTOD \cite{hosseini2020simple}, DoTS \cite{jeon2021domain}, SOLOIST 
 \cite{peng2021soloist}, MinTL \cite{lin2020mintl}, PPTOD \cite{su2021multi}, UBAR \cite{yang2021ubar}, GALAXY \cite{he2022galaxy}, MTTOD \cite{lee2021improving}, BORT \cite{sun2022bort}, Mars \cite{sun2023mars} and SPACE \cite{he2022unified}.

\subsection{Main Results}
As shown in Table \ref{tab:main}, our method TA\&AT achieves new state-of-the-art combined scores on all the datasets in \textbf{w.o. Continual} setting. Even compared with the SOTA SPACE model, performance of our method is comparable, indicating that our proposed methods are competitive for end-to-end task-oriented dialog modeling. Note that based on MTTOD, our method can improve its performance on MultiWOZ 2.0 by 2.23 points (from 107.04 to 109.27), MultiWOZ 2.1 by 1.81 points (from 106.22 to 108.03), MultiWOZ 2.2 by 3.39 points (from 100.2 to 103.59). Note that our model acheives best \textbf{BLEU} in each dataset while keeping other metrics at a high-level, verifying the effectiveness of our method in improving the generation quality.
\subsection{Low-Resource Evaluation}
In order to explore whether our method is equally effective in low-resource scenarios, following the setting of Mars, we tested the performance of the model with 10\%, 20\%, and 50\% number of training sessions, respectively. As shown in Table \ref{tab:lowresource} our method achieves the best in most of data ratio, demonstrating its robustness.

\section{Analysis}

In this section, we first analyze the effectiveness of each auxiliary task and scheduled sampling. Then we discuss some observations from the learning curve in our training process. 
\subsection{Ablation Study}
As shown in Table \ref{tab:ablation}, most auxiliary tasks are effective especially those slot-related ones. Interestingly, we found a phenomenon that removing the response keywords prediction task results in a higher combined score. It seems that this task does not work. We attribute this to the fact that different losses have different learning periods, that is, the learning of $\mathcal{L}_k$ may still be underfitted when the $\mathcal{L}_{sc/st}$ are already overfit (see Section \ref{sec:Curve}), finding an optimal balanced point or determining the best ratio may be a future direction. Besides, such a small increase ($+0.07$) can also be due to randomness. 

In a word, both AT and TA are effective, and the latter is more important, because because it provides a better encoder representation, influencing both understanding
and generation. Compared to most similar baseline MTTOD, our method outperforms it by 2.23, verifying the effectiveness of TA\&AT.

\begin{table}
\large
\centering
\resizebox{0.9\columnwidth}{!}
    {
		{
			\begin{tabular}{l c c c c}
				\hline
				\textbf{Model} &\textbf{Inform} & \textbf{Success} & \textbf{BLEU} & \textbf{Comb}\\
				\hline
				\textbf{TA\&AT} & 93.60 & 83.60 & \textbf{20.67} & 109.27 \\
				 - $\mathcal{L}_{st}$ & 93.10 & 84.50 & 19.79 & 108.59 (-0.68) \\
				 - $\mathcal{L}_{sc}$ & 92.60 & \textbf{84.50} & 20.28 & 108.83 (-0.44) \\ 
                     - $\mathcal{L}_{a}$ & 93.50 & 84.40 & 20.05 & 109.00 (-0.27) \\ 
                     - $\mathcal{L}_{k}$ & \textbf{93.70} & 84.40 & 20.29 & \textbf{109.34} (+0.07) \\ 
				\hdashline
				 w.o. AT & 93.40 & 83.50 & 19.72 & 108.17 (-1.10) \\
                 w.o. TA & 92.90 & 83.30 & 19.76 & 107.86 (-1.41)
 \\
 \hdashline
 MTTOD & 90.99 & 82.58 & 20.25 & 107.04 (-2.23) \\
				\hline
			\end{tabular}
		}
    }
		\caption{Ablation study on E2E results of MultiWOZ 2.0. `w.o. AT' means normal teacher-forcing without any action-tree based scheduled sampling. }
        \label{tab:ablation} 
	\end{table}

\subsection{Learning Curve}\label{sec:Curve}
\begin{figure}
\centering
\includegraphics[width=0.8\columnwidth]{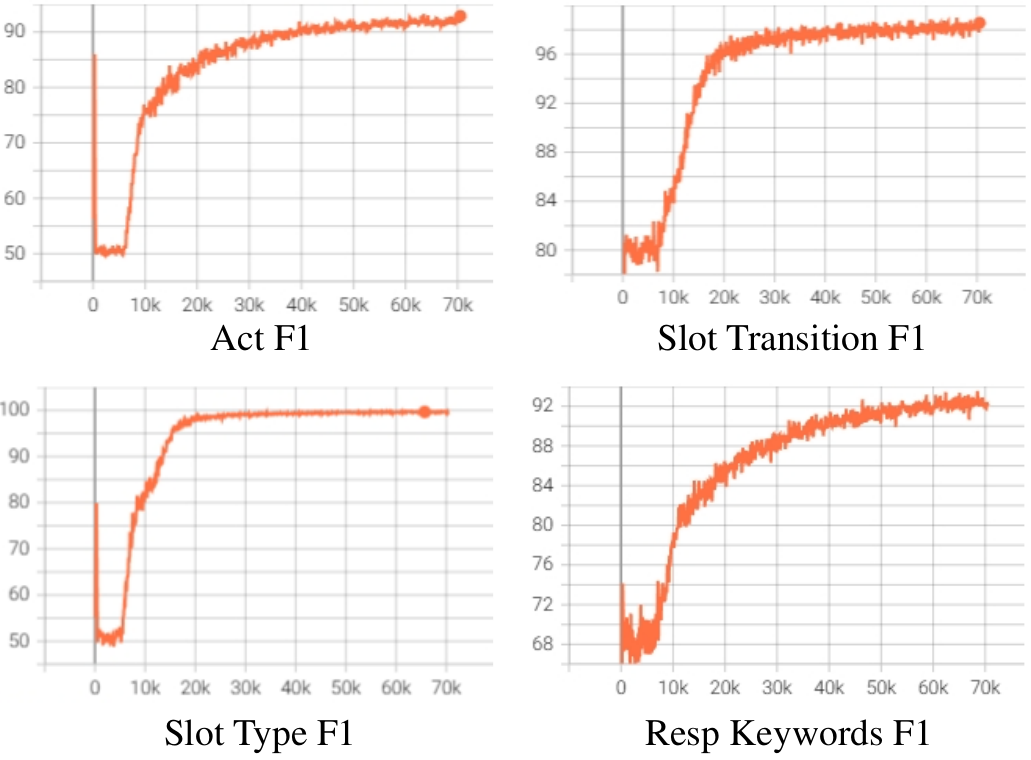} 
\caption{Learning curve for different tasks in training. X-axis represents the number of training steps and Y-axis represents macro F1-score.}
\label{fig:curve}
\end{figure} 
The variation of F1-score corresponding to different tasks during training is shown in Figure \ref{fig:curve}. It can be found that at the very early stage of training, the F1 value is almost unchanged, indicating that the generation loss is primal at this time. In the process of learning the initial generation ability, the hidden state space changes greatly, causing the classifier difficult to train, and it will be relatively easier to predict random/all-1/all-0. As the training progresses, the generation loss decreases and the proportion of auxiliary loss increases, at which time the auxiliary tasks can be optimized. In addition, the tasks related to belief state converge quickly and can reach F1-score above 96, while the tasks related to policy converge slowly and can only reach F1-score around 92. It can be seen that the latter is more difficult than the former, because it requires more planning ability besides understanding.

\subsection{Case Study}
\begin{figure}
\centering
\includegraphics[width=\columnwidth]{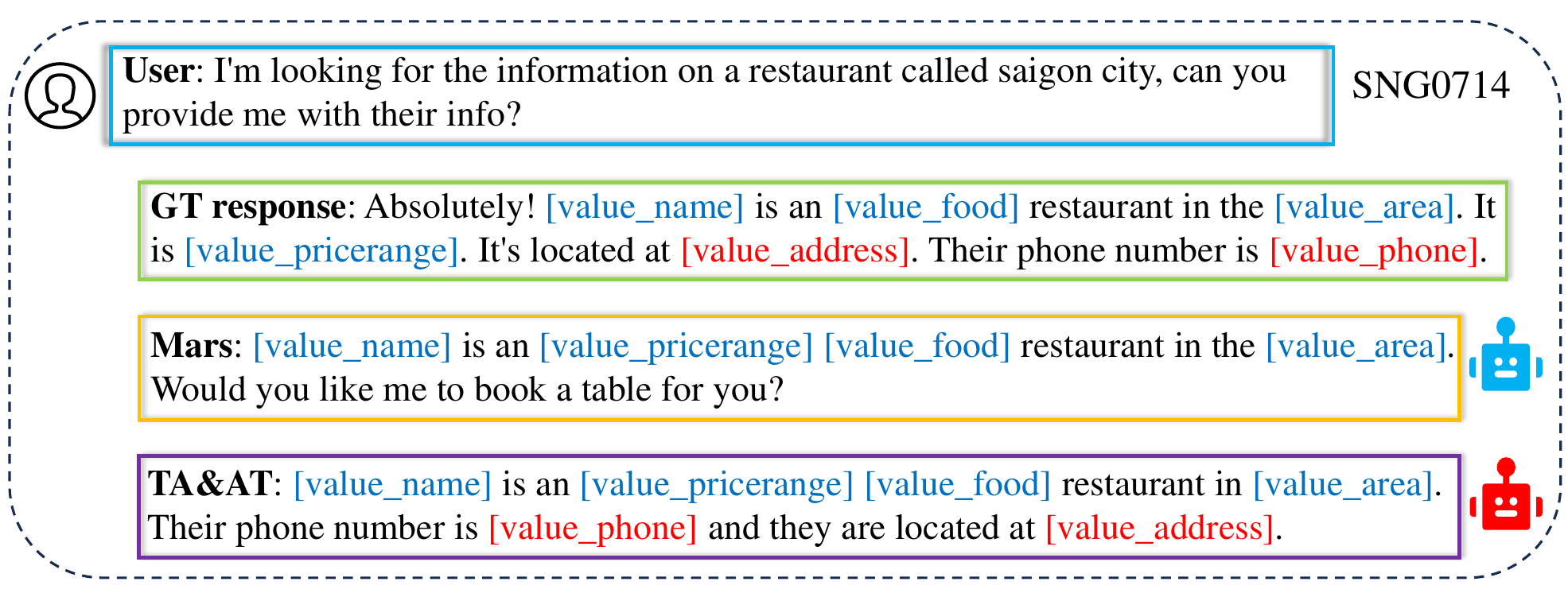} 
\caption{Case Study: Delexicalized responses generated by Mars and TA\&AT on MultiWOZ 2.0 test data. `GT' is short for ground truth.}
\label{fig:case}
\end{figure} 
As shown in Figure \ref{fig:case}, our method can generate more keywords than Mars when the user needs some information, covering all the information contained in the ground truth response and containing no redundant information.

\section{Conclusion}
In this study, we explore the techniques for optimizing task-oriented dialog via turn-level auxiliary tasks and action-tree based scheduled sampling. To address the insufficient utilization of labels and sequence-level error accumulation issues that existing models struggle with, we primarily introduce turn-level multi-task objectives for the encoder module. 
Furthermore, we introduce an action-tree based scheduled sampling technique for the decoder module. Our approach has depicted superior performance on the MultiWOZ dataset series compared to methods without continual pre-training and remains competitive even when benchmarked against methods that adopt pre-training. 

\section*{Acknowledgments}
We thank all the anonymous reviewers for their insightful and valuable comments. This work was supported by National Key R\&D Program of China (NO. 2018AAA0102502) and Independent Research Project of Medical Engineering Laboratory of Chinese PLA General Hospital  (2022SYSZZKY23).

\bibliography{aaai24}

\begin{thebibliography}{33}
\providecommand{\natexlab}[1]{#1}

\bibitem[{Bang, Lee, and Koo(2023)}]{bang2023task}
Bang, N.; Lee, J.; and Koo, M.-W. 2023.
\newblock Task-Optimized Adapters for an End-to-End Task-Oriented Dialogue System.
\newblock \emph{arXiv preprint arXiv:2305.02468}.

\bibitem[{Bengio et~al.(2015)Bengio, Vinyals, Jaitly, and Shazeer}]{bengio2015scheduled}
Bengio, S.; Vinyals, O.; Jaitly, N.; and Shazeer, N. 2015.
\newblock Scheduled sampling for sequence prediction with recurrent neural networks.
\newblock \emph{Advances in neural information processing systems}, 28.

\bibitem[{Budzianowski et~al.(2018)Budzianowski, Wen, Tseng, Casanueva, Ultes, Ramadan, and Ga{\v{s}}i{\'c}}]{budzianowski2018multiwoz}
Budzianowski, P.; Wen, T.-H.; Tseng, B.-H.; Casanueva, I.; Ultes, S.; Ramadan, O.; and Ga{\v{s}}i{\'c}, M. 2018.
\newblock Multiwoz--a large-scale multi-domain wizard-of-oz dataset for task-oriented dialogue modelling.
\newblock \emph{arXiv preprint arXiv:1810.00278}.

\bibitem[{Cholakov and Kolev(2022)}]{cholakov2022efficient}
Cholakov, R.; and Kolev, T. 2022.
\newblock Efficient Task-Oriented Dialogue Systems with Response Selection as an Auxiliary Task.
\newblock \emph{arXiv preprint arXiv:2208.07097}.

\bibitem[{Dong et~al.(2019)Dong, Yang, Wang, Wei, Liu, Wang, Gao, Zhou, and Hon}]{dong2019unified}
Dong, L.; Yang, N.; Wang, W.; Wei, F.; Liu, X.; Wang, Y.; Gao, J.; Zhou, M.; and Hon, H.-W. 2019.
\newblock Unified language model pre-training for natural language understanding and generation.
\newblock \emph{Advances in neural information processing systems}, 32.

\bibitem[{Gu et~al.(2016)Gu, Lu, Li, and Li}]{gu2016incorporating}
Gu, J.; Lu, Z.; Li, H.; and Li, V.~O. 2016.
\newblock Incorporating copying mechanism in sequence-to-sequence learning.
\newblock \emph{arXiv preprint arXiv:1603.06393}.

\bibitem[{He et~al.(2022{\natexlab{a}})He, Dai, Yang, Sun, Huang, Si, and Li}]{he2022unified}
He, W.; Dai, Y.; Yang, M.; Sun, J.; Huang, F.; Si, L.; and Li, Y. 2022{\natexlab{a}}.
\newblock Unified dialog model pre-training for task-oriented dialog understanding and generation.
\newblock In \emph{Proceedings of the 45th International ACM SIGIR Conference on Research and Development in Information Retrieval}, 187--200.

\bibitem[{He et~al.(2022{\natexlab{b}})He, Dai, Zheng, Wu, Cao, Liu, Jiang, Yang, Huang, Si et~al.}]{he2022galaxy}
He, W.; Dai, Y.; Zheng, Y.; Wu, Y.; Cao, Z.; Liu, D.; Jiang, P.; Yang, M.; Huang, F.; Si, L.; et~al. 2022{\natexlab{b}}.
\newblock Galaxy: A generative pre-trained model for task-oriented dialog with semi-supervised learning and explicit policy injection.
\newblock In \emph{Proceedings of the AAAI conference on artificial intelligence}, volume~36, 10749--10757.

\bibitem[{Hochreiter and Schmidhuber(1997)}]{hochreiter1997long}
Hochreiter, S.; and Schmidhuber, J. 1997.
\newblock Long short-term memory.
\newblock \emph{Neural computation}, 9(8): 1735--1780.

\bibitem[{Hosseini-Asl et~al.(2020)Hosseini-Asl, McCann, Wu, Yavuz, and Socher}]{hosseini2020simple}
Hosseini-Asl, E.; McCann, B.; Wu, C.-S.; Yavuz, S.; and Socher, R. 2020.
\newblock A simple language model for task-oriented dialogue.
\newblock \emph{Advances in Neural Information Processing Systems}, 33: 20179--20191.

\bibitem[{Jeon and Lee(2021)}]{jeon2021domain}
Jeon, H.; and Lee, G.~G. 2021.
\newblock Domain state tracking for a simplified dialogue system.
\newblock \emph{arXiv preprint arXiv:2103.06648}.

\bibitem[{Kalchbrenner, Grefenstette, and Blunsom(2014)}]{kalchbrenner2014convolutional}
Kalchbrenner, N.; Grefenstette, E.; and Blunsom, P. 2014.
\newblock A convolutional neural network for modelling sentences.
\newblock \emph{arXiv preprint arXiv:1404.2188}.

\bibitem[{Kim et~al.(2019)Kim, Yang, Kim, and Lee}]{kim2019efficient}
Kim, S.; Yang, S.; Kim, G.; and Lee, S.-W. 2019.
\newblock Efficient dialogue state tracking by selectively overwriting memory.
\newblock \emph{arXiv preprint arXiv:1911.03906}.

\bibitem[{Kulh{\'a}nek et~al.(2021)Kulh{\'a}nek, Hude{\v{c}}ek, Nekvinda, and Du{\v{s}}ek}]{kulhanek2021augpt}
Kulh{\'a}nek, J.; Hude{\v{c}}ek, V.; Nekvinda, T.; and Du{\v{s}}ek, O. 2021.
\newblock AuGPT: Auxiliary tasks and data augmentation for end-to-end dialogue with pre-trained language models.
\newblock \emph{arXiv preprint arXiv:2102.05126}.

\bibitem[{Lee(2021)}]{lee2021improving}
Lee, Y. 2021.
\newblock Improving end-to-end task-oriented dialog system with a simple auxiliary task.
\newblock In \emph{Findings of the Association for Computational Linguistics: EMNLP 2021}, 1296--1303.

\bibitem[{Lei et~al.(2018)Lei, Jin, Kan, Ren, He, and Yin}]{lei2018sequicity}
Lei, W.; Jin, X.; Kan, M.-Y.; Ren, Z.; He, X.; and Yin, D. 2018.
\newblock Sequicity: Simplifying task-oriented dialogue systems with single sequence-to-sequence architectures.
\newblock In \emph{Proceedings of the 56th Annual Meeting of the Association for Computational Linguistics (Volume 1: Long Papers)}, 1437--1447.

\bibitem[{Li et~al.(2021)Li, Zhang, Fei, Feng, and Zhou}]{li2021conversations}
Li, Z.; Zhang, J.; Fei, Z.; Feng, Y.; and Zhou, J. 2021.
\newblock Conversations are not flat: Modeling the dynamic information flow across dialogue utterances.
\newblock \emph{arXiv preprint arXiv:2106.02227}.

\bibitem[{Lin et~al.(2020)Lin, Madotto, Winata, and Fung}]{lin2020mintl}
Lin, Z.; Madotto, A.; Winata, G.~I.; and Fung, P. 2020.
\newblock Mintl: Minimalist transfer learning for task-oriented dialogue systems.
\newblock \emph{arXiv preprint arXiv:2009.12005}.

\bibitem[{Loshchilov and Hutter(2017)}]{loshchilov2017decoupled}
Loshchilov, I.; and Hutter, F. 2017.
\newblock Decoupled weight decay regularization.
\newblock \emph{arXiv preprint arXiv:1711.05101}.

\bibitem[{Peng et~al.(2021)Peng, Li, Li, Shayandeh, Liden, and Gao}]{peng2021soloist}
Peng, B.; Li, C.; Li, J.; Shayandeh, S.; Liden, L.; and Gao, J. 2021.
\newblock Soloist: Building task bots at scale with transfer learning and machine teaching.
\newblock \emph{Transactions of the Association for Computational Linguistics}, 9: 807--824.

\bibitem[{Radford et~al.(2018)Radford, Narasimhan, Salimans, Sutskever et~al.}]{radford2018improving}
Radford, A.; Narasimhan, K.; Salimans, T.; Sutskever, I.; et~al. 2018.
\newblock Improving language understanding by generative pre-training.

\bibitem[{Raffel et~al.(2020)Raffel, Shazeer, Roberts, Lee, Narang, Matena, Zhou, Li, and Liu}]{raffel2020exploring}
Raffel, C.; Shazeer, N.; Roberts, A.; Lee, K.; Narang, S.; Matena, M.; Zhou, Y.; Li, W.; and Liu, P.~J. 2020.
\newblock Exploring the limits of transfer learning with a unified text-to-text transformer.
\newblock \emph{The Journal of Machine Learning Research}, 21(1): 5485--5551.

\bibitem[{Su et~al.(2021)Su, Shu, Mansimov, Gupta, Cai, Lai, and Zhang}]{su2021multi}
Su, Y.; Shu, L.; Mansimov, E.; Gupta, A.; Cai, D.; Lai, Y.-A.; and Zhang, Y. 2021.
\newblock Multi-task pre-training for plug-and-play task-oriented dialogue system.
\newblock \emph{arXiv preprint arXiv:2109.14739}.

\bibitem[{Sun et~al.(2022)Sun, Bao, Wu, and He}]{sun2022bort}
Sun, H.; Bao, J.; Wu, Y.; and He, X. 2022.
\newblock BORT: Back and denoising reconstruction for end-to-end task-oriented dialog.
\newblock \emph{arXiv preprint arXiv:2205.02471}.

\bibitem[{Sun et~al.(2023)Sun, Bao, Wu, and He}]{sun2023mars}
Sun, H.; Bao, J.; Wu, Y.; and He, X. 2023.
\newblock Mars: Modeling Context \& State Representations with Contrastive Learning for End-to-End Task-Oriented Dialog.
\newblock In \emph{Findings of the Association for Computational Linguistics: ACL 2023}, 11139--11160.

\bibitem[{Takanobu et~al.(2020)Takanobu, Zhu, Li, Peng, Gao, and Huang}]{takanobu2020your}
Takanobu, R.; Zhu, Q.; Li, J.; Peng, B.; Gao, J.; and Huang, M. 2020.
\newblock Is your goal-oriented dialog model performing really well? empirical analysis of system-wise evaluation.
\newblock \emph{arXiv preprint arXiv:2005.07362}.

\bibitem[{Wen et~al.(2016)Wen, Vandyke, Mrksic, Gasic, Rojas-Barahona, Su, Ultes, and Young}]{wen2016network}
Wen, T.-H.; Vandyke, D.; Mrksic, N.; Gasic, M.; Rojas-Barahona, L.~M.; Su, P.-H.; Ultes, S.; and Young, S. 2016.
\newblock A network-based end-to-end trainable task-oriented dialogue system.
\newblock \emph{arXiv preprint arXiv:1604.04562}.

\bibitem[{Wolf et~al.(2020)Wolf, Debut, Sanh, Chaumond, Delangue, Moi, Cistac, Rault, Louf, Funtowicz et~al.}]{wolf2020transformers}
Wolf, T.; Debut, L.; Sanh, V.; Chaumond, J.; Delangue, C.; Moi, A.; Cistac, P.; Rault, T.; Louf, R.; Funtowicz, M.; et~al. 2020.
\newblock Transformers: State-of-the-art natural language processing.
\newblock In \emph{Proceedings of the 2020 conference on empirical methods in natural language processing: system demonstrations}, 38--45.

\bibitem[{Yang, Li, and Quan(2021)}]{yang2021ubar}
Yang, Y.; Li, Y.; and Quan, X. 2021.
\newblock UBAR: Towards fully end-to-end task-oriented dialog system with GPT-2.
\newblock In \emph{Proceedings of the AAAI Conference on Artificial Intelligence}, volume~35, 14230--14238.

\bibitem[{Zhang and Shasha(1989)}]{zhang1989simple}
Zhang, K.; and Shasha, D. 1989.
\newblock Simple fast algorithms for the editing distance between trees and related problems.
\newblock \emph{SIAM journal on computing}, 18(6): 1245--1262.

\bibitem[{Zhang et~al.(2019)Zhang, Feng, Meng, You, and Liu}]{zhang2019bridging}
Zhang, W.; Feng, Y.; Meng, F.; You, D.; and Liu, Q. 2019.
\newblock Bridging the gap between training and inference for neural machine translation.
\newblock \emph{arXiv preprint arXiv:1906.02448}.

\bibitem[{Zhang et~al.(2020)Zhang, Ou, Wang, and Feng}]{zhang2020probabilistic}
Zhang, Y.; Ou, Z.; Wang, H.; and Feng, J. 2020.
\newblock A probabilistic end-to-end task-oriented dialog model with latent belief states towards semi-supervised learning.
\newblock \emph{arXiv preprint arXiv:2009.08115}.

\bibitem[{Zhang, Ou, and Yu(2020)}]{zhang2020task}
Zhang, Y.; Ou, Z.; and Yu, Z. 2020.
\newblock Task-oriented dialog systems that consider multiple appropriate responses under the same context.
\newblock In \emph{Proceedings of the AAAI Conference on Artificial Intelligence}, volume~34, 9604--9611.

\end{thebibliography}

\end{document}